\documentclass{article}
\usepackage{arxiv}

\usepackage{graphicx}
\usepackage{booktabs}
\usepackage{multirow}
\usepackage{amsmath}
\usepackage{url}
\usepackage{enumitem}
\usepackage{hyperref}
\usepackage{caption}
\usepackage{subcaption}
\usepackage{float}

\title{A Multimodal Multi-Agent Framework for Radiology Report Generation}

\author{
  Ziruo Yi \\
  University of North Texas \\
  \texttt{ziruoyi@my.unt.edu}
  \And
  Ting Xiao \\
  University of North Texas \\
  \texttt{ting.xiao@unt.edu}
  \And
  Mark V. Albert \\
   University of North Texas \\
  \texttt{mark.albert@unt.edu}
}

\begin{document}

\maketitle

\begin{abstract}
Radiology report generation (RRG) aims to automatically produce diagnostic reports from medical images, with the potential to enhance clinical workflows and reduce radiologists' workload. While recent approaches leveraging multimodal large language models (MLLMs) and retrieval-augmented generation (RAG) have achieved strong results, they continue to face challenges such as factual inconsistency, hallucination, and cross-modal misalignment. We propose a multimodal multi-agent framework for RRG that aligns with the stepwise clinical reasoning workflow, where task-specific agents handle retrieval, draft generation, visual analysis, refinement, and synthesis. Experimental results demonstrate that our approach outperforms a strong baseline in both automatic metrics and LLM-based evaluations, producing more accurate, structured, and interpretable reports. This work highlights the potential of clinically aligned multi-agent frameworks to support explainable and trustworthy clinical AI applications.
\end{abstract}


\section{Introduction}

Radiology plays an essential role in modern healthcare, supporting diagnosis, treatment planning, and outcome prediction. It involves diverse data sources such as chest X-ray images, laboratory results, and clinical notes. Among multimodal tasks in radiology, radiology report generation (RRG) is particularly impactful as it directly supports clinical workflows and decision-making. Recent studies further emphasize the importance of RRG as it aligns closely with radiologists’ documentation~\cite{yildirim2024multimodal, yi2025survey}. RRG typically involves two key modalities: chest X-ray images that provide visual evidence of patient conditions, and corresponding radiology reports that capture clinical details in natural language. However, increasing demand for radiological exams and a shortage of radiologists have led to delays, forcing clinicians to make critical decisions without radiological guidance, which may result in errors or conclusions that differ from those of experienced radiologists~\cite{mcdonald2015effects, petinaux2011accuracy}.

With advancements in artificial intelligence (AI), computer vision (CV), and natural language processing (NLP), multimodal learning has emerged as a powerful paradigm for integrating and analyzing diverse data sources~\cite{huang2021makes, waqas2024multimodal}. Recently, large language models (LLMs) and large vision models (LVMs), including GPT-4V~\cite{achiam2023gpt}, LLaMA~3~\cite{meta2024introducing}, and DALL·E~3~\cite{openai2023dalle3}, have gained substantial attention. Building on these advances, multimodal large language models (MLLMs) have demonstrated strong performance on tasks like image captioning~\cite{li2023blip} and visual-language dialogue~\cite{huang2023sparkles}. In the medical domain, MLLMs such as Med-PaLM 2~\cite{singhal2025toward} and LLaVA-Med~\cite{li2023llava} have made notable progress in pharmaceutical research~\cite{grisoni2023chemical} and clinical support~\cite{ali2023using}. For RRG, MLLMs integrate visual and textual information to generate detailed and clinically accurate reports~\cite{liu2023radiology, danu2023generation}, facilitating structured documentation and supporting clinical decision-making. These applications extend the capabilities of radiologists, reduce workload, and assist less experienced clinicians. Despite this progress, existing MLLM-based approaches face key limitations in RRG. First, although MLLMs can process visual inputs effectively, they often struggle when essential information is textual or requires cross-modal reasoning. Second, most existing systems lack a unified architecture that flexibly integrates techniques such as prompt engineering. This limits their adaptability to new requirements in RRG. Third, current methods typically lack intermediate validation or refinement stages, making them prone to factual inconsistencies and hallucinations.

Retrieval-augmented generation (RAG)~\cite{gao2023retrieval, qu2024alleviating, qu2024look} has emerged as a promising method to enhance the factual accuracy of medical MLLMs. By integrating external and reliable sources, RAG enriches the model’s knowledge and supports more grounded generation. It has been applied to tasks such as visual question answering (VQA)~\cite{yuan2023ramm} and report generation~\cite{kumar2024improving, tao2024memory}.
However, applying RAG to medical MLLMs introduces several new challenges. While retrieving too few contexts may miss relevant information, retrieving too many can introduce noise and redundancy, making it harder for the model to identify relevant content and ultimately degrading output quality.

To address the limitations of existing MLLM and RAG approaches, we propose a multimodal multi-agent framework for RRG that decomposes the task into five specialized agents. Our framework combines RAG with a collaborative multi-agent system in which specialized agents jointly process and integrate visual and textual information. It begins with a Retrieval Agent that selects top-$k$ similar reports for a given chest X-ray image. These retrieved examples are passed to a Draft Agent, which generates an initial version of the report. A Refiner Agent then extracts key findings from both the draft and the retrieved context to highlight essential diagnostic information. In parallel, a Vision Agent produces a description summarizing visual observations from the chest X-ray image. Finally, the Synthesis Agent integrates the outputs from the vision, retrieval, and refiner agents to generate the final report. This agent-driven workflow follows the stepwise clinical reasoning process, assigning distinct roles to agents in a modular and interpretable manner. The contributions of our work are: 
(1) We propose a clinically aligned multi-agent framework for RRG, enabling modular collaboration across task-specific agents and incorporating RAG to enhance factuality and controllability.
(2) We conduct extensive experiments demonstrating that our method consistently outperforms a strong single-agent baseline across both automatic metrics and LLM-based evaluations.

\section{Related Work}

\textbf{MLLMs for RRG.} MLLMs have recently emerged as a promising solution for automating RRG~\cite{ji2024vision, liu2024bootstrapping, wang2024trrg, zhou2024large, lu2023effectively}. Models such as R2GenGPT~\cite{wang2023r2gengpt}, XrayGPT~\cite{thawkar2023xraygpt}, and MAIRA-1~\cite{hyland2023maira} combine visual encoders (e.g., Swin Transformer~\cite{liu2021swin}, MedCLIP~\cite{wang2022medclip}) with LLMs (e.g., LLaMA~\cite{meta2024introducing}, Vicuna~\cite{chiang2023vicuna}) to align visual features with textual representations, demonstrating strong performance on benchmark datasets. Despite their success, these models still suffer from key limitations including factual inconsistency~\cite{xia2024cares, su2024conflictbank}, hallucination~\cite{liu2024survey}, and catastrophic forgetting~\cite{zhai2023investigating, khan2023importance}. These issues are particularly critical in RRG, where factual accuracy and reliability are essential for clinical applications.

\textbf{Retrieval-Augmented Generation.} RAG has been widely adopted to enhance factual accuracy by incorporating contextual information from external datasets~\cite{gao2023retrieval, sun2024surf}. It has been applied to RRG to reduce hallucinations and enhance content relevance~\cite{ranjit2023retrieval, sun2024fact, xia2024mmed, liang2024optimizing, bernardi2024report}. However, current RAG techniques face critical challenges: the number and quality of retrieved contexts, and the risk of over-reliance on these references, both of which may degrade model performance or introduce factual errors~\cite{xia2024rule}. Moreover, existing RAG methods often retrieve and process text and image information independently, limiting their ability to perform integrated multimodal reasoning~\cite{han2025mdocagent}. These limitations are particularly problematic in RRG, which depends on fine-grained alignment between retrieved knowledge and visual evidence.

\textbf{Multi-Agent Systems.} Multi-agent systems have gained increasing attention in NLP and healthcare AI~\cite{wang2024survey, yue2024ct, ke2024enhancing, wei2024medco, tang2023medagents, smit2023we}. They assign different tasks to specialized agents, which collaborate to accomplish complex goals that single models often struggle with. Preliminary attempts have explored multi-agent paradigms for RRG~\cite{zeng2024enhancing, alam2025towards}, showing promising results. However, applying multi-agent systems to multimodal tasks introduces new challenges. In particular, simply combining outputs from isolated vision and text agents often fails to capture the cross-modal relationships required for accurate interpretation. Moreover, aligning agent interactions with domain-specific workflows such as stepwise clinical reasoning remains a key challenge in current systems. To address these limitations, we propose a multimodal multi-agent framework aligned with stepwise clinical reasoning, where task-specific agents handle retrieval, draft generation, refinement, visual analysis, and synthesis in a modular and interpretable manner.

\section{Method}

\begin{figure}[!htbp]
  \centering
  \includegraphics[width=\linewidth]{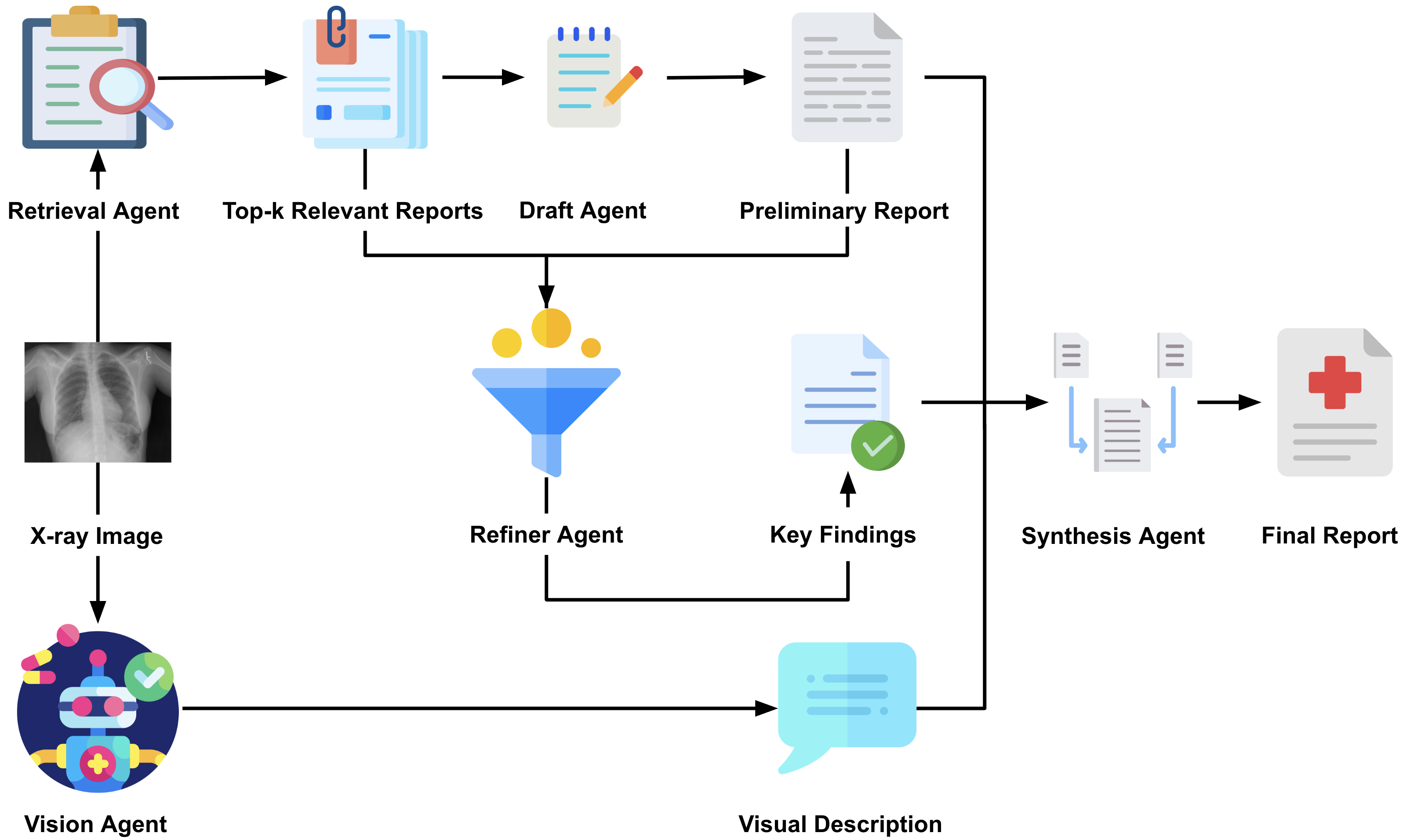} 
  \caption{Overview of our proposed multi-agent framework for automated RRG. The system decomposes the task into five agents: (1) Retrieval Agent selects top-k similar reports. (2) Draft Agent generates a preliminary report from the retrieved texts. (3) Refiner Agent distills key clinical findings. (4) Vision Agent generates a visual description of the image. (5) Synthesis Agent integrates these outputs to produce the final report.}
  \label{fig:framework}
\end{figure}

We propose a modular multi-agent framework for RRG, designed to emulate the clinical workflow by combining case retrieval, visual interpretation, and structured textual synthesis. Given a chest X-ray image, the system sequentially activates five specialized agents: a Retrieval Agent, a Draft Agent, a Refiner Agent, a Vision Agent, and a Synthesis Agent. Each agent fulfills a distinct functional role and operates independently, using either task-specific prompts (for LLM/VLM agents) or embedding-based retrieval (for the retrieval module). As shown in Figure~\ref{fig:framework}, agents communicate via structured intermediate outputs that progressively refine radiological observations into a final, coherent impression. This design promotes factual accuracy, improves interpretability, and helps ensure consistent report generation.

\vspace{0.5em}
\vspace{0.25em}

\subsection{Retrieval Agent}

The Retrieval Agent performs cross-modal retrieval by identifying prior radiology reports that are semantically similar to a given chest X-ray image. Inspired by the design of CLIP~\cite{radford2021learning}, the agent encodes the input image into a visual embedding and compares it against report embeddings using cosine similarity. The top-$k$ most similar reports are selected based on similarity scores, where $k$ is a predefined parameter balancing retrieval coverage and efficiency. These reports provide relevant diagnostic context, such as clinical findings and report style, which serve as guidance for downstream generation.

\subsection{Draft Agent}

The Draft Agent composes a preliminary radiology report by synthesizing information from the top-k reports selected by the Retrieval Agent. Inspired by how radiologists review similar prior cases, the agent identifies shared clinical findings and prioritizes medically relevant observations. It then organizes this information into a clinically focused report that reflects an initial diagnostic impression. This intermediate output provides a structured textual basis for downstream processing by later agents.

\subsection{Refiner Agent}

The Refiner Agent distills key clinical findings from the outputs of the Draft and Retrieval Agents. It is designed to identify clinically important observations that are consistently supported by the input. Unlike the Draft Agent, which generates a broad preliminary report, the Refiner Agent focuses solely on findings-level content. It receives both the preliminary report and the original retrieved reports as input, and outputs a concise, single-paragraph summary containing the most essential findings. To ensure factuality, the agent enforces retrieval-grounded constraints: every sentence must be clearly supported by the input, with no speculation or paraphrasing beyond factual rewriting. The output provides a structured clinical signal for downstream synthesis.

\subsection{Vision Agent}

The Vision Agent generates a visual description of the chest X-ray image to complement textual information from previous agents. It uses a medical MLLM to generate image-grounded descriptions based on visible observations in the input image. The output is a caption describing key chest regions, such as the lungs and mediastinum. The agent is designed to avoid unclear statements and irrelevant content, ensuring that the caption is grounded in visible evidence and written in a radiology report style. This step introduces visual cues from the input image to support the final synthesis.

\subsection{Synthesis Agent}

The Synthesis Agent produces the final radiology report by integrating a preliminary report, critical findings, and a visual caption from the previous agents. To ensure both factual consistency and stylistic coherence, the final report includes only observations explicitly supported by the textual or visual inputs. The agent is designed to avoid unsupported findings and unnecessary rewriting, while preserving the core clinical content from each input and combining them in a logically consistent manner. This final step concludes the multi-agent pipeline by generating a clinically grounded and well-structured radiology report.

\section{Experiments}

In this section, we evaluate our multimodal multi-agent framework by addressing the following questions: 
(1) Does the multi-agent design improve the clinical accuracy of generated radiology reports compared to the baseline?  
(2)~Does each agent play a meaningful role in the generation process? 
(3) How does the framework enhance the overall quality of the generated reports?

\subsection{Experimental Setup}

\textbf{Implementation Details.}  
Our framework consists of five agents: a retrieval agent, a draft agent, a refiner agent, a vision agent, and a synthesis agent.  
We follow the retrieval setup of RULE~\cite{xia2024rule} to construct the retrieval agent, which fine-tunes CLIP on MIMIC-CXR using contrastive learning to adapt to the medical domain.  
LLaVA-Med 1.5 (7B)~\cite{li2023llava} is used as the backbone for the vision agent, while GPT-4o~\cite{achiam2023gpt} powers the draft, refiner, and synthesis agents. The retrieval agent selects the top-$k$ most similar reports ($k=5$ by default), which are then passed to the draft agent as input.  

\textbf{Datasets.} 
We utilize two publicly available chest X-ray datasets: MIMIC-CXR~\cite{johnson2019mimic} and IU X-ray~\cite{demner2016preparing}. 
We fine-tune the retrieval agent using 3,000 image–report pairs from MIMIC-CXR, a large-scale dataset of chest X-rays with associated radiology reports. 
For evaluation, we use the IU X-ray dataset, which includes chest radiographs and corresponding diagnostic reports. 
Following the data split from~\cite{liu2024factual}, the IU X-ray dataset contains 2,068 training and 590 test image–report pairs after filtering. 
We use the training set to construct the retrieval database and the test set to evaluate the performance of our framework.

\textbf{Evaluation Metrics.} The performance of our multi-agent framework is evaluated using standard metrics for text generation, including BLEU~\cite{papineni2002bleu}, ROUGE-1, ROUGE-2, ROUGE-L~\cite{lin2004rouge}, and BERTScore~\cite{zhang2019bertscore}. These metrics focus on surface-level similarity between generated and reference impressions, primarily based on lexical or token overlap. To complement these automatic metrics, we adopt the LLM-as-a-Judge paradigm~\cite{zheng2023judging} and employ Claude 3 Opus~\cite{claude3haiku2024}, a state-of-the-art LLM developed by Anthropic, to assess both the semantic accuracy and clinical relevance of the generated reports.

\subsection{Results}

In this section, we present a comprehensive evaluation of our multi-agent framework on the IU X-ray dataset, comparing it against a single-agent baseline using LLaVA-Med that simulates a radiologist working without access to prior reports or clinical cues.

\subsubsection{Quantitative Analysis}

\textbf{Standard Metrics.} We evaluate the performance of our framework using standard metrics, including BLEU, ROUGE, METEOR, and BERTScore. The results of this evaluation are listed in Table~\ref{tab:quant-comparison}. Our multi-agent framework achieves a BLEU score of 0.0466, significantly outperforming LLaVA-Med at 0.0036. ROUGE-1, ROUGE-2, and ROUGE-L scores increase from 0.2398, 0.0278, and 0.1537 to 0.3652, 0.1292, and 0.2471 respectively, demonstrating consistent gains across all ROUGE metrics. For METEOR, the score rises to 0.3618 from 0.1437, indicating better lexical diversity and content coverage. On BERTScore, the framework achieves 0.8819 compared to 0.8617 for the baseline, suggesting stronger semantic alignment between generated and reference texts. These results indicate that the multi-agent design substantially improves both textual quality and semantic coherence in RRG.

\begin{table}[!htbp]
\centering
\caption{Quantitative performance comparison between our multi-agent framework and a single MLLM.}
\begin{tabular}{lcccccc}
\toprule
Model & BLEU & ROUGE-1 & ROUGE-2 & ROUGE-L & METEOR & BERTScore \\
\midrule
Llava-Med & 0.0036 & 0.2398 & 0.0278 & 0.1537 & 0.1437 & 0.8617 \\
\textbf{Ours} & \textbf{0.0466} & \textbf{0.3652} & \textbf{0.1292} & \textbf{0.2471} & \textbf{0.3618} & \textbf{0.8819} \\
\bottomrule
\end{tabular}
\label{tab:quant-comparison}
\end{table}
\vspace{0.75em}

\textbf{LLM-as-a-Judge.}
To complement standard metrics, we further assess the clinical and linguistic quality of the generated reports using Claude 3 Opus~\cite{claude3haiku2024}, focusing on five key aspects: coverage of key findings, consistency with original reports, diagnostic accuracy, stylistic alignment, and conciseness. Each aspect is rated from 1 to 10, with higher scores indicating better quality. As summarized in Table~\ref{tab:llmjudge}, our multi-agent framework outperforms LLaVA-Med on four of the five evaluation dimensions. It achieves a diagnostic accuracy score of 8.26 compared to 7.78, demonstrating stronger clinical reasoning.
Our framework scores 8.16 in style alignment and 7.26 in conciseness, surpassing the baseline scores of 7.98 and 6.98. These improvements reflect stronger alignment with clinical report writing standards. Key finding coverage also improves from 5.86 to 6.36, showing a clearer presentation of clinically relevant information. Although LLaVA-Med slightly leads in consistency (6.94 vs. 6.74), the overall results highlight the effectiveness of our multi-agent design in enhancing clinical reliability and writing quality.

\begin{table}[!htbp]
  \centering
  \caption{Qualitative performance comparison between our multi-agent framework and a single MLLM.}
  \label{tab:llmjudge}
  \begin{tabular}{lccccc}
    \toprule
    Model & Findings & Consistency & Diagnosis & Style & Conciseness \\
    \midrule
    LLaVA-Med & 5.86 & \textbf{6.94} & 7.78 & 7.98 & 6.98 \\
    Ours      & \textbf{6.36} & 6.74 & \textbf{8.26} & \textbf{8.16} & \textbf{7.26} \\
    \bottomrule
  \end{tabular}
\end{table}
\vspace{0.75em}

\subsubsection{Qualitative Analysis}

\begin{figure}[!htbp]
  \centering
  \includegraphics[width=\linewidth]{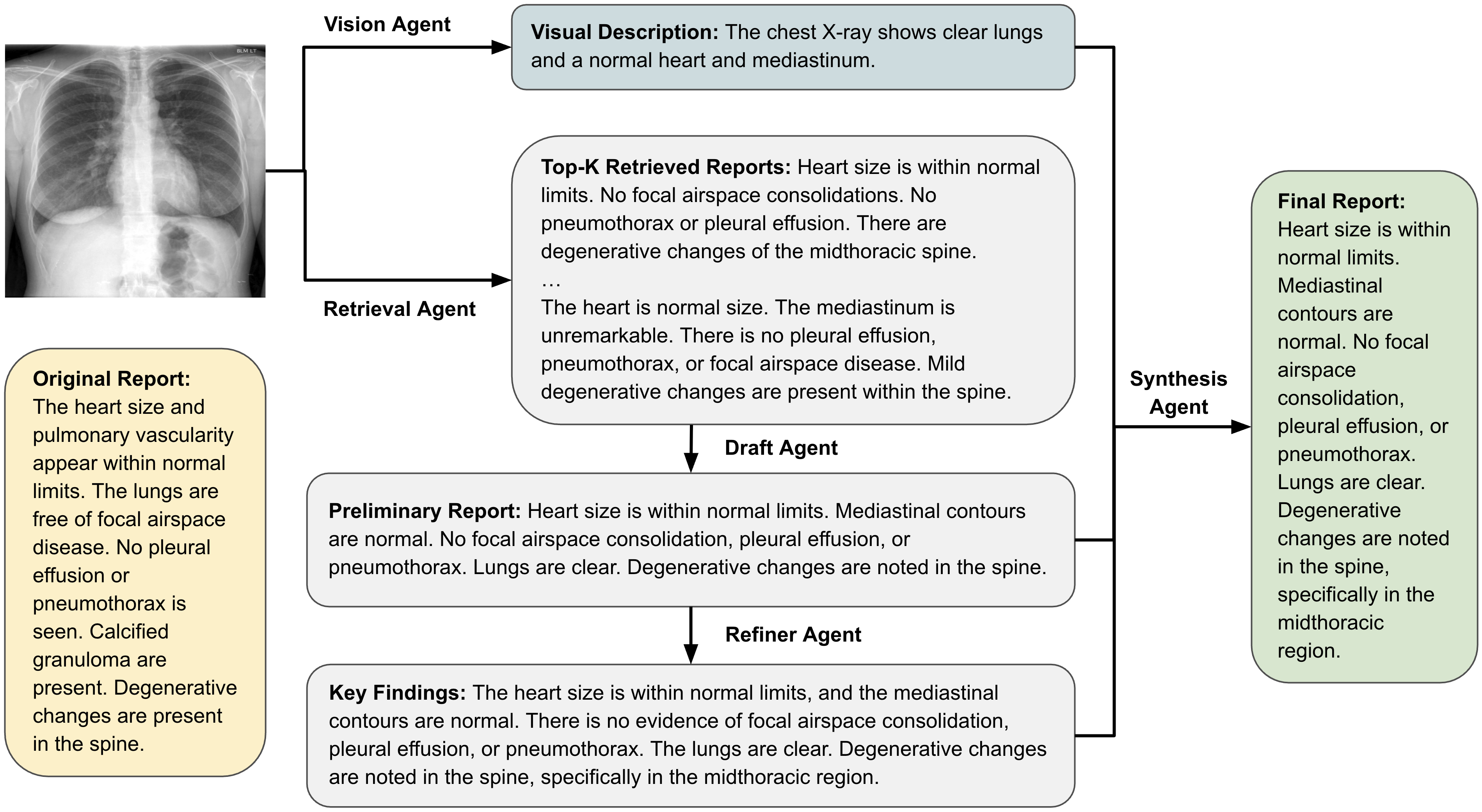}
  \caption{
  A case study showing that retrieval and key findings help overcome the limitations of a vision-only agent.
  }
  \label{fig:case_study}
\end{figure}

Figure~\ref{fig:case_study} presents a representative example comparing the report generated by the Vision Agent only with that of our full multi-agent framework. This case highlights the benefit of incorporating retrieved reports and extracted key findings through our multi-agent pipeline. The vision-only output, while stylistically reasonable, lacks specificity and misses important observations. In contrast, the multi-agent output offers a more complete and clinically aligned summary. It follows terminology and structure commonly seen in prior reports, such as the inclusion of “pleural effusion” and “Degenerative changes are noted in the spine,” better aligning with the original report. These improvements are enabled by structured agent collaboration: the Retrieval Agent supplies relevant contextual examples, the Refiner Agent extracts key clinical findings, and the Synthesis Agent combines them with the visual caption into a structured report. The final output is more concise, better organized, and clinically reliable, illustrating how retrieval grounding and agent collaboration lead to higher-quality reports than those produced by a vision-only agent.

\subsection{Discussion}

Our results show that the proposed multi-agent framework consistently enhances RRG across both standard automatic metrics and LLM-based clinical assessment. By assigning each agent a specific function such as retrieval, abstraction, refinement, visual captioning, and synthesis, the framework introduces a clearer structure and separation of responsibilities. This modularity enables more controllable and interpretable generation, allowing each agent to focus on a distinct aspect of clinical reasoning or stylistic consistency. The generated reports are more complete and stylistically aligned. They also show stronger clinical grounding and better adherence to radiology reporting conventions.

While our framework demonstrates notable improvements, we observe a slight drop in consistency compared to the baseline. Similar findings have been reported in prior work~\cite{zeng2024enhancing}, where RAG can introduce redundant or irrelevant content, reducing overall coherence. In our case, the combination of retrieved reports, refined findings, and visual descriptions introduces additional complexity, which can affect the overall consistency of the final output. Despite this limitation, our experiments demonstrate that the multi-agent framework improves clinical accuracy and report quality compared to a strong single-agent baseline, as validated by both automatic and LLM-based evaluations. The case study further highlights the distinct contributions of individual agents in enhancing factuality and structure. Future work includes a more systematic investigation, particularly through agent-level ablation.

\section{Conclusion}

We present a multimodal multi-agent framework for RRG that breaks down the task into specialized agents for retrieval, draft, refinement, visual interpretation, and synthesis. Our approach follows the diagnostic reasoning process and outperforms a strong single-agent MLLM baseline, as demonstrated by both automatic metrics and LLM-based evaluations. Through the collaboration among agents, the framework produces reports that are more clinically grounded, coherent, and stylistically aligned. This modular design offers a generalizable method for other multimodal medical tasks requiring diagnostic reasoning and clinical precision.

\bibliographystyle{unsrt}
\bibliography{template}

\begin{thebibliography}{10}

\bibitem{yildirim2024multimodal}
Nur Yildirim, Hannah Richardson, Maria~Teodora Wetscherek, Junaid Bajwa, Joseph Jacob, Mark~Ames Pinnock, Stephen Harris, Daniel Coelho De~Castro, Shruthi Bannur, Stephanie Hyland, et~al.
\newblock Multimodal healthcare ai: identifying and designing clinically relevant vision-language applications for radiology.
\newblock In {\em Proceedings of the 2024 CHI Conference on Human Factors in Computing Systems}, pages 1--22, 2024.

\bibitem{yi2025survey}
Ziruo Yi, Ting Xiao, and Mark~V Albert.
\newblock A survey on multimodal large language models in radiology for report generation and visual question answering.
\newblock {\em Information}, 16(2):136, 2025.

\bibitem{mcdonald2015effects}
Robert~J McDonald, Kara~M Schwartz, Laurence~J Eckel, Felix~E Diehn, Christopher~H Hunt, Brian~J Bartholmai, Bradley~J Erickson, and David~F Kallmes.
\newblock The effects of changes in utilization and technological advancements of cross-sectional imaging on radiologist workload.
\newblock {\em Academic radiology}, 22(9):1191--1198, 2015.

\bibitem{petinaux2011accuracy}
Bruno Petinaux, Rahul Bhat, Keith Boniface, and Jaime Aristizabal.
\newblock Accuracy of radiographic readings in the emergency department.
\newblock {\em The American journal of emergency medicine}, 29(1):18--25, 2011.

\bibitem{huang2021makes}
Yu~Huang, Chenzhuang Du, Zihui Xue, Xuanyao Chen, Hang Zhao, and Longbo Huang.
\newblock What makes multi-modal learning better than single (provably).
\newblock {\em Advances in Neural Information Processing Systems}, 34:10944--10956, 2021.

\bibitem{waqas2024multimodal}
Asim Waqas, Aakash Tripathi, Ravi~P Ramachandran, Paul~A Stewart, and Ghulam Rasool.
\newblock Multimodal data integration for oncology in the era of deep neural networks: a review.
\newblock {\em Frontiers in Artificial Intelligence}, 7:1408843, 2024.

\bibitem{achiam2023gpt}
Josh Achiam, Steven Adler, Sandhini Agarwal, Lama Ahmad, Ilge Akkaya, Florencia~Leoni Aleman, Diogo Almeida, Janko Altenschmidt, Sam Altman, Shyamal Anadkat, et~al.
\newblock Gpt-4 technical report.
\newblock {\em arXiv preprint arXiv:2303.08774}, 2023.

\bibitem{meta2024introducing}
AI~Meta.
\newblock Introducing meta llama 3: The most capable openly available llm to date.
\newblock {\em Meta AI}, 2024.

\bibitem{openai2023dalle3}
OpenAI.
\newblock D{ALL-E}3, 2023.
\newblock \url{https://openai.com/index/dall-e-3/}.

\bibitem{li2023blip}
Junnan Li, Dongxu Li, Silvio Savarese, and Steven Hoi.
\newblock Blip-2: Bootstrapping language-image pre-training with frozen image encoders and large language models.
\newblock In {\em International conference on machine learning}, pages 19730--19742. PMLR, 2023.

\bibitem{huang2023sparkles}
Yupan Huang, Zaiqiao Meng, Fangyu Liu, Yixuan Su, Nigel Collier, and Yutong Lu.
\newblock Sparkles: Unlocking chats across multiple images for multimodal instruction-following models.
\newblock {\em arXiv preprint arXiv:2308.16463}, 2023.

\bibitem{singhal2025toward}
Karan Singhal, Tao Tu, Juraj Gottweis, Rory Sayres, Ellery Wulczyn, Mohamed Amin, Le~Hou, Kevin Clark, Stephen~R Pfohl, Heather Cole-Lewis, et~al.
\newblock Toward expert-level medical question answering with large language models.
\newblock {\em Nature Medicine}, pages 1--8, 2025.

\bibitem{li2023llava}
Chunyuan Li, Cliff Wong, Sheng Zhang, Naoto Usuyama, Haotian Liu, Jianwei Yang, Tristan Naumann, Hoifung Poon, and Jianfeng Gao.
\newblock Llava-med: Training a large language-and-vision assistant for biomedicine in one day.
\newblock {\em Advances in Neural Information Processing Systems}, 36:28541--28564, 2023.

\bibitem{grisoni2023chemical}
Francesca Grisoni.
\newblock Chemical language models for de novo drug design: Challenges and opportunities.
\newblock {\em Current Opinion in Structural Biology}, 79:102527, 2023.

\bibitem{ali2023using}
Stephen~R Ali, Thomas~D Dobbs, Hayley~A Hutchings, and Iain~S Whitaker.
\newblock Using chatgpt to write patient clinic letters.
\newblock {\em The Lancet Digital Health}, 5(4):e179--e181, 2023.

\bibitem{liu2023radiology}
Zhengliang Liu, Yiwei Li, Peng Shu, Aoxiao Zhong, Longtao Yang, Chao Ju, Zihao Wu, Chong Ma, Jie Luo, Cheng Chen, et~al.
\newblock Radiology-llama2: Best-in-class large language model for radiology.
\newblock {\em arXiv preprint arXiv:2309.06419}, 2023.

\bibitem{danu2023generation}
Manuela~Daniela Danu, George Marica, Sanjeev~Kumar Karn, Bogdan Georgescu, Awais Mansoor, Florin Ghesu, Lucian~Mihai Itu, Constantin Suciu, Sasa Grbic, Oladimeji Farri, et~al.
\newblock Generation of radiology findings in chest x-ray by leveraging collaborative knowledge.
\newblock {\em Procedia Computer Science}, 221:1102--1109, 2023.

\bibitem{gao2023retrieval}
Yunfan Gao, Yun Xiong, Xinyu Gao, Kangxiang Jia, Jinliu Pan, Yuxi Bi, Yixin Dai, Jiawei Sun, Haofen Wang, and Haofen Wang.
\newblock Retrieval-augmented generation for large language models: A survey.
\newblock {\em arXiv preprint arXiv:2312.10997}, 2:1, 2023.

\bibitem{qu2024alleviating}
Xiaoye Qu, Qiyuan Chen, Wei Wei, Jishuo Sun, and Jianfeng Dong.
\newblock Alleviating hallucination in large vision-language models with active retrieval augmentation.
\newblock {\em arXiv preprint arXiv:2408.00555}, 2024.

\bibitem{qu2024look}
Xiaoye Qu, Jiashuo Sun, Wei Wei, and Yu~Cheng.
\newblock Look, compare, decide: Alleviating hallucination in large vision-language models via multi-view multi-path reasoning.
\newblock {\em arXiv preprint arXiv:2408.17150}, 2024.

\bibitem{yuan2023ramm}
Zheng Yuan, Qiao Jin, Chuanqi Tan, Zhengyun Zhao, Hongyi Yuan, Fei Huang, and Songfang Huang.
\newblock Ramm: Retrieval-augmented biomedical visual question answering with multi-modal pre-training.
\newblock In {\em Proceedings of the 31st ACM International Conference on Multimedia}, pages 547--556, 2023.

\bibitem{kumar2024improving}
Yogesh Kumar and Pekka Marttinen.
\newblock Improving medical multi-modal contrastive learning with expert annotations.
\newblock In {\em European Conference on Computer Vision}, pages 468--486. Springer, 2024.

\bibitem{tao2024memory}
Yitian Tao, Liyan Ma, Jing Yu, and Han Zhang.
\newblock Memory-based cross-modal semantic alignment network for radiology report generation.
\newblock {\em IEEE Journal of Biomedical and Health Informatics}, 2024.

\bibitem{ji2024vision}
Jia Ji, Yongshuai Hou, Xinyu Chen, Youcheng Pan, and Yang Xiang.
\newblock Vision-language model for generating textual descriptions from clinical images: Model development and validation study.
\newblock {\em JMIR Formative Research}, 8:e32690, 2024.

\bibitem{liu2024bootstrapping}
Chang Liu, Yuanhe Tian, Weidong Chen, Yan Song, and Yongdong Zhang.
\newblock Bootstrapping large language models for radiology report generation.
\newblock In {\em Proceedings of the AAAI Conference on Artificial Intelligence}, volume~38, pages 18635--18643, 2024.

\bibitem{wang2024trrg}
Yuhao Wang, Chao Hao, Yawen Cui, Xinqi Su, Weicheng Xie, Tao Tan, and Zitong Yu.
\newblock Trrg: Towards truthful radiology report generation with cross-modal disease clue enhanced large language model.
\newblock {\em arXiv preprint arXiv:2408.12141}, 2024.

\bibitem{zhou2024large}
Zijian Zhou, Miaojing Shi, Meng Wei, Oluwatosin Alabi, Zijie Yue, and Tom Vercauteren.
\newblock Large model driven radiology report generation with clinical quality reinforcement learning.
\newblock {\em arXiv preprint arXiv:2403.06728}, 2024.

\bibitem{lu2023effectively}
Yuzhe Lu, Sungmin Hong, Yash Shah, and Panpan Xu.
\newblock Effectively fine-tune to improve large multimodal models for radiology report generation.
\newblock {\em arXiv preprint arXiv:2312.01504}, 2023.

\bibitem{wang2023r2gengpt}
Zhanyu Wang, Lingqiao Liu, Lei Wang, and Luping Zhou.
\newblock R2gengpt: Radiology report generation with frozen llms.
\newblock {\em Meta-Radiology}, 1(3):100033, 2023.

\bibitem{thawkar2023xraygpt}
Omkar Thawkar, Abdelrahman Shaker, Sahal~Shaji Mullappilly, Hisham Cholakkal, Rao~Muhammad Anwer, Salman Khan, Jorma Laaksonen, and Fahad~Shahbaz Khan.
\newblock Xraygpt: Chest radiographs summarization using medical vision-language models.
\newblock {\em arXiv preprint arXiv:2306.07971}, 2023.

\bibitem{hyland2023maira}
Stephanie~L Hyland, Shruthi Bannur, Kenza Bouzid, Daniel~C Castro, Mercy Ranjit, Anton Schwaighofer, Fernando P{\'e}rez-Garc{\'\i}a, Valentina Salvatelli, Shaury Srivastav, Anja Thieme, et~al.
\newblock Maira-1: A specialised large multimodal model for radiology report generation.
\newblock {\em arXiv preprint arXiv:2311.13668}, 2023.

\bibitem{liu2021swin}
Ze~Liu, Yutong Lin, Yue Cao, Han Hu, Yixuan Wei, Zheng Zhang, Stephen Lin, and Baining Guo.
\newblock Swin transformer: Hierarchical vision transformer using shifted windows.
\newblock In {\em Proceedings of the IEEE/CVF international conference on computer vision}, pages 10012--10022, 2021.

\bibitem{wang2022medclip}
Zifeng Wang, Zhenbang Wu, Dinesh Agarwal, and Jimeng Sun.
\newblock Medclip: Contrastive learning from unpaired medical images and text.
\newblock In {\em Proceedings of the Conference on Empirical Methods in Natural Language Processing. Conference on Empirical Methods in Natural Language Processing}, volume 2022, page 3876, 2022.

\bibitem{chiang2023vicuna}
Wei-Lin Chiang, Zhuohan Li, Zi~Lin, Ying Sheng, Zhanghao Wu, Hao Zhang, Lianmin Zheng, Siyuan Zhuang, Yonghao Zhuang, Joseph~E Gonzalez, et~al.
\newblock Vicuna: An open-source chatbot impressing gpt-4 with 90\%* chatgpt quality, march 2023.
\newblock {\em URL https://lmsys. org/blog/2023-03-30-vicuna}, 3(5), 2023.

\bibitem{xia2024cares}
Peng Xia, Ze~Chen, Juanxi Tian, Yangrui Gong, Ruibo Hou, Yue Xu, Zhenbang Wu, Zhiyuan Fan, Yiyang Zhou, Kangyu Zhu, et~al.
\newblock Cares: A comprehensive benchmark of trustworthiness in medical vision language models.
\newblock {\em Advances in Neural Information Processing Systems}, 37:140334--140365, 2024.

\bibitem{su2024conflictbank}
Zhaochen Su, Jun Zhang, Xiaoye Qu, Tong Zhu, Yanshu Li, Jiashuo Sun, Juntao Li, Min Zhang, and Yu~Cheng.
\newblock Conflictbank: A benchmark for evaluating the influence of knowledge conflicts in llm.
\newblock {\em arXiv preprint arXiv:2408.12076}, 2024.

\bibitem{liu2024survey}
Hanchao Liu, Wenyuan Xue, Yifei Chen, Dapeng Chen, Xiutian Zhao, Ke~Wang, Liping Hou, Rongjun Li, and Wei Peng.
\newblock A survey on hallucination in large vision-language models.
\newblock {\em arXiv preprint arXiv:2402.00253}, 2024.

\bibitem{zhai2023investigating}
Yuexiang Zhai, Shengbang Tong, Xiao Li, Mu~Cai, Qing Qu, Yong~Jae Lee, and Yi~Ma.
\newblock Investigating the catastrophic forgetting in multimodal large language models.
\newblock {\em arXiv preprint arXiv:2309.10313}, 2023.

\bibitem{khan2023importance}
Hikmat Khan, Nidhal~C Bouaynaya, and Ghulam Rasool.
\newblock The importance of robust features in mitigating catastrophic forgetting.
\newblock In {\em 2023 IEEE Symposium on Computers and Communications (ISCC)}, pages 752--757. IEEE, 2023.

\bibitem{sun2024surf}
Jiashuo Sun, Jihai Zhang, Yucheng Zhou, Zhaochen Su, Xiaoye Qu, and Yu~Cheng.
\newblock Surf: Teaching large vision-language models to selectively utilize retrieved information.
\newblock {\em arXiv preprint arXiv:2409.14083}, 2024.

\bibitem{ranjit2023retrieval}
Mercy Ranjit, Gopinath Ganapathy, Ranjit Manuel, and Tanuja Ganu.
\newblock Retrieval augmented chest x-ray report generation using openai gpt models.
\newblock In {\em Machine Learning for Healthcare Conference}, pages 650--666. PMLR, 2023.

\bibitem{sun2024fact}
Liwen Sun, James Zhao, Megan Han, and Chenyan Xiong.
\newblock Fact-aware multimodal retrieval augmentation for accurate medical radiology report generation.
\newblock {\em arXiv preprint arXiv:2407.15268}, 2024.

\bibitem{xia2024mmed}
Peng Xia, Kangyu Zhu, Haoran Li, Tianze Wang, Weijia Shi, Sheng Wang, Linjun Zhang, James Zou, and Huaxiu Yao.
\newblock Mmed-rag: Versatile multimodal rag system for medical vision language models.
\newblock {\em arXiv preprint arXiv:2410.13085}, 2024.

\bibitem{liang2024optimizing}
Siting Liang, Pablo S{\'a}nchez, and Daniel Sonntag.
\newblock Optimizing relation extraction in medical texts through active learning: A comparative analysis of trade-offs.
\newblock In {\em Proceedings of the 1st Workshop on Uncertainty-Aware NLP (UncertaiNLP 2024)}, pages 23--34, 2024.

\bibitem{bernardi2024report}
Mario~Luca Bernardi and Marta Cimitile.
\newblock Report generation from x-ray imaging by retrieval-augmented generation and improved image-text matching.
\newblock In {\em 2024 International Joint Conference on Neural Networks (IJCNN)}, pages 1--8. IEEE, 2024.

\bibitem{xia2024rule}
Peng Xia, Kangyu Zhu, Haoran Li, Hongtu Zhu, Yun Li, Gang Li, Linjun Zhang, and Huaxiu Yao.
\newblock Rule: Reliable multimodal rag for factuality in medical vision language models.
\newblock In {\em Proceedings of the 2024 Conference on Empirical Methods in Natural Language Processing}, pages 1081--1093, 2024.

\bibitem{han2025mdocagent}
Siwei Han, Peng Xia, Ruiyi Zhang, Tong Sun, Yun Li, Hongtu Zhu, and Huaxiu Yao.
\newblock Mdocagent: A multi-modal multi-agent framework for document understanding.
\newblock {\em arXiv preprint arXiv:2503.13964}, 2025.

\bibitem{wang2024survey}
Lei Wang, Chen Ma, Xueyang Feng, Zeyu Zhang, Hao Yang, Jingsen Zhang, Zhiyuan Chen, Jiakai Tang, Xu~Chen, Yankai Lin, et~al.
\newblock A survey on large language model based autonomous agents.
\newblock {\em Frontiers of Computer Science}, 18(6):186345, 2024.

\bibitem{yue2024ct}
Ling Yue and Tianfan Fu.
\newblock Ct-agent: Clinical trial multi-agent with large language model-based reasoning.
\newblock {\em arXiv e-prints}, pages arXiv--2404, 2024.

\bibitem{ke2024enhancing}
Yu~He Ke, Rui Yang, Sui~An Lie, Taylor Xin~Yi Lim, Hairil~Rizal Abdullah, Daniel Shu~Wei Ting, and Nan Liu.
\newblock Enhancing diagnostic accuracy through multi-agent conversations: using large language models to mitigate cognitive bias.
\newblock {\em arXiv preprint arXiv:2401.14589}, 2024.

\bibitem{wei2024medco}
Hao Wei, Jianing Qiu, Haibao Yu, and Wu~Yuan.
\newblock Medco: Medical education copilots based on a multi-agent framework.
\newblock {\em arXiv preprint arXiv:2408.12496}, 2024.

\bibitem{tang2023medagents}
Xiangru Tang, Anni Zou, Zhuosheng Zhang, Ziming Li, Yilun Zhao, Xingyao Zhang, Arman Cohan, and Mark Gerstein.
\newblock Medagents: Large language models as collaborators for zero-shot medical reasoning.
\newblock {\em arXiv preprint arXiv:2311.10537}, 2023.

\bibitem{smit2023we}
Andries~Petrus Smit, Paul Duckworth, Nathan Grinsztajn, Kale-ab Tessera, Thomas~D Barrett, and Arnu Pretorius.
\newblock Are we going mad? benchmarking multi-agent debate between language models for medical q\&a.
\newblock In {\em Deep Generative Models for Health Workshop NeurIPS 2023}, 2023.

\bibitem{zeng2024enhancing}
Fang Zeng, Zhiliang Lyu, Quanzheng Li, and Xiang Li.
\newblock Enhancing llms for impression generation in radiology reports through a multi-agent system.
\newblock {\em arXiv preprint arXiv:2412.06828}, 2024.

\bibitem{alam2025towards}
Hasan Md~Tusfiqur Alam, Devansh Srivastav, Md~Abdul Kadir, and Daniel Sonntag.
\newblock Towards interpretable radiology report generation via concept bottlenecks using a multi-agentic rag.
\newblock In {\em European Conference on Information Retrieval}, pages 201--209. Springer, 2025.

\bibitem{radford2021learning}
Alec Radford, Jong~Wook Kim, Chris Hallacy, Aditya Ramesh, Gabriel Goh, Sandhini Agarwal, Girish Sastry, Amanda Askell, Pamela Mishkin, Jack Clark, et~al.
\newblock Learning transferable visual models from natural language supervision.
\newblock In {\em International conference on machine learning}, pages 8748--8763. PmLR, 2021.

\bibitem{johnson2019mimic}
Alistair~EW Johnson, Tom~J Pollard, Nathaniel~R Greenbaum, Matthew~P Lungren, Chih-ying Deng, Yifan Peng, Zhiyong Lu, Roger~G Mark, Seth~J Berkowitz, and Steven Horng.
\newblock Mimic-cxr-jpg, a large publicly available database of labeled chest radiographs.
\newblock {\em arXiv preprint arXiv:1901.07042}, 2019.

\bibitem{demner2016preparing}
Dina Demner-Fushman, Marc~D Kohli, Marc~B Rosenman, Sonya~E Shooshan, Laritza Rodriguez, Sameer Antani, George~R Thoma, and Clement~J McDonald.
\newblock Preparing a collection of radiology examinations for distribution and retrieval.
\newblock {\em Journal of the American Medical Informatics Association}, 23(2):304--310, 2016.

\bibitem{liu2024factual}
Kang Liu, Zhuoqi Ma, Mengmeng Liu, Zhicheng Jiao, Xiaolu Kang, Qiguang Miao, and Kun Xie.
\newblock Factual serialization enhancement: A key innovation for chest x-ray report generation.
\newblock {\em arXiv preprint arXiv:2405.09586}, 2024.

\bibitem{papineni2002bleu}
Kishore Papineni, Salim Roukos, Todd Ward, and Wei-Jing Zhu.
\newblock Bleu: a method for automatic evaluation of machine translation.
\newblock In {\em Proceedings of the 40th annual meeting of the Association for Computational Linguistics}, pages 311--318, 2002.

\bibitem{lin2004rouge}
Chin-Yew Lin.
\newblock Rouge: A package for automatic evaluation of summaries.
\newblock In {\em Text summarization branches out}, pages 74--81, 2004.

\bibitem{zhang2019bertscore}
Tianyi Zhang, Varsha Kishore, Felix Wu, Kilian~Q Weinberger, and Yoav Artzi.
\newblock Bertscore: Evaluating text generation with bert.
\newblock {\em arXiv preprint arXiv:1904.09675}, 2019.

\bibitem{zheng2023judging}
Lianmin Zheng, Wei-Lin Chiang, Ying Sheng, Siyuan Zhuang, Zhanghao Wu, Yonghao Zhuang, Zi~Lin, Zhuohan Li, Dacheng Li, Eric Xing, et~al.
\newblock Judging llm-as-a-judge with mt-bench and chatbot arena.
\newblock {\em Advances in Neural Information Processing Systems}, 36:46595--46623, 2023.

\bibitem{claude3haiku2024}
Anthropic.
\newblock Claude 3 haiku: Our fastest model yet, 2024.
\newblock \url{https://www.anthropic.com/news/claude-3-haiku}.

\end{thebibliography}

\end{document}